\def\year{2019}\relax
\pdfoutput=1
\documentclass[letterpaper]{article} %DO NOT CHANGE THIS
\usepackage{aaai19}  %Required
\usepackage{times}  %Required
\usepackage{helvet}  %Required
\usepackage{courier}  %Required
\usepackage{url}  %Required
\usepackage{graphicx}  %Required
\usepackage{booktabs}
\usepackage{multicol}
\usepackage{amsmath}
\usepackage{amssymb}
\usepackage{algorithm}
\usepackage{multirow}
\usepackage{stfloats}
\usepackage[noend]{algpseudocode}
\usepackage{caption,subcaption}

\makeatletter
\def\Bstate{\State\hskip-\ALG@thistlm}
\makeatother

\begin{document}
\title{Practical Text Classification With Large Pre-Trained Language Models}
%\author{Anonymous}
\author{
Neel Kant\\
University of California, Berkeley\\
\texttt{kantneel@berkeley.edu}
\And
Raul Puri\\
NVIDIA, Santa Clara CA\\
\texttt{raulp@nvidia.com}
\And
Nikolai Yakovenko\\
NVIDIA, Santa Clara CA\\
\texttt{nyakovenko@nvidia.com}
\And
Bryan Catanzaro\\
NVIDIA, Santa Clara CA\\
\texttt{bcatanzaro@nvidia.com}
}
\maketitle

\begin{abstract}
Multi-emotion sentiment classification is a natural language processing (NLP) problem with valuable use cases on real-world data. We demonstrate that large-scale unsupervised language modeling combined with finetuning offers a practical solution to this task on difficult datasets, including those with label class imbalance and domain-specific context. By training an attention-based Transformer network \protect\cite{Transformer2017} on 40GB of text (Amazon reviews) \protect\cite{McAuley2015} and fine-tuning on the training set, our model achieves a 0.69 F1 score on the SemEval Task 1:E-c multidimensional emotion classification problem \protect\cite{SemEval2018Task1}, based on the Plutchik wheel of emotions \protect\cite{Plutchik1979}. These results are competitive with state of the art models, including strong F1 scores on difficult (emotion) categories such as Fear (0.73), Disgust (0.77) and Anger (0.78), as well as competitive results on rare categories such as Anticipation (0.42) and Surprise (0.37). Furthermore, we demonstrate our application on a 
real world text classification task. We create a narrowly collected text dataset of real tweets on several topics, and show that our finetuned model outperforms general purpose commercially available APIs for sentiment and multidimensional emotion classification on this dataset by a significant margin.
We also perform a variety of additional studies, investigating properties of deep learning architectures, datasets and algorithms for achieving practical multidimensional sentiment classification. Overall, we find that unsupervised language modeling and finetuning is a simple framework for achieving high quality results on real-world sentiment classification. 

\end{abstract}

\section{Introduction}
Recent work has shown that language models -- both RNN variants like the multiplicative LSTM (mLSTM) \protect\cite{Krause2016}, as well as the attention-based Transformer network \protect\cite{Transformer2017} -- can be trained efficiently over very large datasets, and that the resulting models can be transferred to downstream language understanding problems, often matching or exceeding the previous state of the art approaches on academic datasets. However, how well do these models perform on practical text classification problems, with real world data?

In this work, we train both mLSTM and Transformer language models on a large 40GB text dataset \protect\cite{McAuley2015}, then transfer those models to two text classification problems: binary sentiment (including Neutral labels), and multidimensional emotion classification based on the Plutchik wheel of emotions \protect\cite{Plutchik1979}. We examine our performance on these tasks, both against large academic datasets, and on an original text dataset that we compiled from social media messages about several specific topics, such as video games. 

We demonstrate that our approach matches the state of the art on the academic datasets without domain-specific training and without excessive hyper-parameter tuning. Meanwhile on the social media dataset, our approach outperforms commercially available APIs by significant margins, even when those models are re-calibrated to the test set. 

Furthermore, we notice that 1) the Transformer model generally out-performs the mLSTM model, especially when fine-tuning on multidimensional emotion classification, and 2) fine-tuning the model significantly improves performance on the emotion tasks, both for the mLSTM and the Transformer model. We suggest that our approach creates models with good generalization to increasingly difficult text classification problems, and we offer ablation studies to demonstrate that effect. 

It is difficult to fit a single model for text classification across domains, due to unknown words, specialized context, colloquial language, and other differences between domains. For example, words such as \textbf{war} and \textbf{sick} are not necessarily negative in the context of video games, which are significantly represented in our dataset. By training a language model across a large text dataset, we expose our model to many contexts. Perhaps a small amount of downstream transfer is enough to choose the right context features for emotion classification in the appropriate setting. 

Our work shows that unsupervised language modeling combined with finetuning offers a practical solution to specialized text classification problems, including those with large category class imbalance, and significant human label disagreement.

% ------------------------------------
% ------------------------------------
% Proposed Problem Statement
% ------------------------------------
% ------------------------------------

\section{Background}
Supervised learning is difficult to apply to NLP problems because labels are expensive. Following \protect\cite{Radford2017}, \protect\cite{Radford2018} and \protect\cite{Dai2015}, we train unsupervised text models on large amounts of unlabelled text data, and transfer the model features to small supervised text problems. The supervised text classification problem used for transfer is binary sentiment on the Stanford Sentiment Treebank (SST) \protect\cite{SST-Socher:2013}.

Some of these binary text examples are subtle. Prior works show that unsupervised language models can learn nuanced features of text, such as word ordering and double negation, just from the underlying task of next-word prediction. However, while this includes difficult examples, it does not necessarily represent sentiment on practical text problems.
\begin{itemize}
\itemsep0em
    \item The source material (professionally written movie reviews) does not include colloquial language.
    \item The dataset excludes Neutral sentiment texts and those with weak directional sentiment.
    \item The dataset does not include dimensions of sentiment apart from Positive and Negative.
\end{itemize}

\begin{table*}[!bp]
\caption{Difficult video game tweets.}
\label{table:video-game-tweets}
\makebox[\textwidth]{
\resizebox{\textwidth}{!}{
\begin{tabular}{p{12cm}|p{1cm}p{1cm}p{1cm}p{1cm}|p{1cm}|p{1cm}}
\toprule
\multirow{2}{*}{\hspace{5cm} Tweet} & Watson & \multirow{2}{*}{Sad} & \multirow{2}{*}{Joy} & \multirow{2}{*}{Fear} & GCL & Ours   \\
& Binary & & & & Binary & Binary \\
\midrule
Encouraging collaboration among players in Sea of \textbf{Thieves} \textless url\textgreater & -0.302 & 0.229 & 0.194 & 0.150 & -0.80 & Pos \\
got my first \textbf{kill} on Fortnite all by myself I'm geeked \textless emoji\textgreater  perioddddd. & -0.847 & 0.003 & 0.666 & 0.225 & +0.60 & Neu \\
Far \textbf{Cry} 5 "\textbf{Lost} On Mars" Gameplay Walkthrough - DLC2: \textless url\textgreater via @YouTube & -0.909 & 0.047 & 0.015 & 0.873 & +0.00 & Neu \\
NEW SUBMACHINE \textbf{GUN} IS \textbf{INSANE}! | Fortnite Best Moments 39 (Fortnite Funny \textbf{Fails} \& WTF Moments) \textless url\textgreater & -0.936 & 0.821 & 0.178 & 0.056 & -0.10 & Pos \\
\bottomrule
\end{tabular}
}
}
\end{table*}
\paragraph{Plutchik's Wheel of Emotions}
We focus our multi-dimension emotion classification on Plutchik's wheel of emotions \protect\cite{Plutchik1979}. This taxonomy, in use since 1979, aims to classify human emotions as a combination of four dualities: Joy - Sadness, Anger - Fear, Trust - Disgust, and Surprise - Anticipation. According to the \textit{basic emotion model} \protect\cite{Ekman2013AnAF}, while humans experience hundreds of emotions, some emotions are more fundamental than others. 

\begin{figure}[t!] 
  \begin{center}
    \includegraphics[width=\columnwidth]{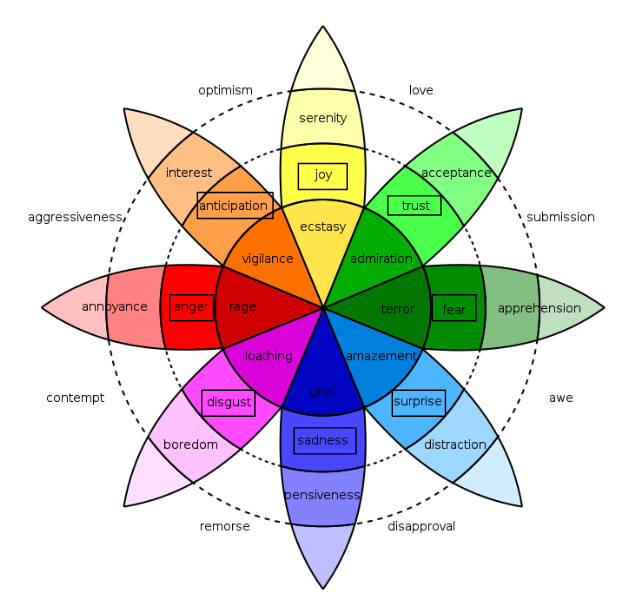}
    \caption{Plutchik's wheel of emotions \protect\cite{Plutchik1979}.}
    \label{fig:Plutchik_wheel}
  \end{center}
\end{figure}

The commercial general purpose emotion classification API that we compare against, IBM's Watson\footnote{https://www.ibm.com/watson/services/natural-language-understanding/}, offers classification scores for the Joy, Sadness, Fear, Disgust and Anger emotions -- all present in Plutchik's wheel (Fig. \ref{fig:Plutchik_wheel}).

\paragraph{SemEval Multidimension Emotion Dataset}

The SemEval Task 1:E-c problem \protect\cite{SemEval2018Task1} offers a training set of 6,857 tweets, with binary labels for the eight Plutchik categories, plus Optimism, Pessimism, and Love. This dataset was created through a process of text selection and human labeling. We show our results on this dataset and compare it to the current state of the art performance.

While it is not possible to report rater agreement on these categories for the compilation of the dataset, the authors note that 2 out of 7 raters had to agree for a positive label to be applied, as requiring larger agreement caused a scarcity of labels for some categories. This indicates that some of the categories had significant rater disagreement between the human raters. The dataset also included a substantial degree of label class imbalance, with some categories like Anger (37\%), Disgust (38\%), Joy (36\%) and Sadness (29\%) represented often in the dataset, while others like Trust (5\%) and Surprise (5\%) present much less frequently (Fig.\ref{table:class-imbalance}). This class imbalance and human rater disagreement is not uncommon for real world text classification problems\footnote{We submitted the SemEval training set for re-labeling using our rater instructions. See Fig.\ref{table:rater-agreement} for an estimate of rater disagreement over the SemEval training set.}.

\begin{table*}[!t]
\centering
\caption{Label class balance (as percent) for SemEval and company tweet datasets.}
\label{table:class-imbalance}
\makebox[\textwidth]{
\resizebox{0.85\textwidth}{!}{%
\begin{tabular}{cccccccccc|c}
\toprule
& \textit{Size} & Anger & Anticipation & Disgust & Fear & Joy & Sad & Surprise & Trust & Ave/None \\
\midrule
SemEval & \textit{6,858} & \textbf{37.2} & 14.3 & \textbf{38.0} & \textbf{18.2} & \textbf{36.2} & \textbf{29.4} & 5.3 & 5.2 & \textbf{23.0}/2.9 \\
(Random) & \textit{4,021} & 7.8 & \textbf{14.7} & 5.2 & 1.7 & 21.9 & 3.4 & 4.3 & 6.6 & 8.2/\textbf{52.1} \\
(Active) & \textit{5,024} & 22.0 & 10.2 & 12.3 & 5.6 & 19.7 & 6.3 & \textbf{7.1} & 6.5 & 11.2/35.6 \\
(All) & \textit{13,326} & 11.7 & 12.9 & 6.8 & 2.9 & 20.6 & 4.2 & 5.0 & \textbf{7.6} & 8.9/47.0\\
\bottomrule
\end{tabular}%
}
}
\end{table*}

\paragraph{Company Tweet Dataset}
In addition to the SemEval tweet dataset, we wanted to see how our model would perform on a similar but domain-specific task: Plutchik emotion classification on tweets relevant to a particular company. We collected tweets on a variety of topics, including:
\begin{itemize}
    \itemsep0em
    \item Video game tweets
    \item Tweets about the company stock
\end{itemize}

We submitted the first batch of 4,000 tweets to human raters on the FigureEight\footnote{https://www.figure-eight.com/} platform, with rules similar to those used by SemEval, which also used the FigureEight platform for human labeling. Specifically, we verified that raters passed our golden set (answering 70\% of test questions correctly). We applied positive labels for each category where 2 out of 5 raters agreed. This is slightly less permissive than the 2 out of 7 raters used by SemEval, because we did not have a budget for 7 raters per tweet.

After the first pass, we noticed that random sampling led to some categories being severely under-sampled, below 5\% of tweets.
Thus we employed a bootstrapping technique to pre-classify tweets by category using our current model, and choose tweets with more likely emotion tweets for classification. See \textbf{Active Learning} section for details. We also sampled 5,000 tweets balanced by source category, since video game tweets have much more emotion, thus dominated the bootstrapped selections. 

Henceforth, we refer to the combined company tweets dataset consisting of: 
\begin{itemize}
    \itemsep0em
    \item 4,021 random tweets
    \item 5,024 tweets selected for higher emotion content
    \item 4,281 tweets selected for source category balance
\end{itemize}

% Quick table about rater disagreement
% NOTE: *Much* of the rater emotion agreement is about "no emotion" -- would be even lower by category -- but no report by category I can easily find.
\begin{table}[ht]
\caption{Inter-rater agreement. Humans don't always agree, even on binary sentiment.}
\label{table:rater-agreement}
\makebox[\columnwidth][l]{
%\makebox{\textwidth}{
%\resizebox{\textwidth}{!}{
\resizebox{\columnwidth}{!}{
\begin{tabular}{cccc}
\toprule
\multirow{2}{*}{Dataset} & \multirow{2}{*}{Judgments} & Binary & Plutchik \\
&  & (3 choices)  & (8 choices) \\
\midrule
SemEval & 20,514 & 77.3\% & 61.1\% \\
Company (random) & 20,005  & 80.7\% & 67.3\% \\
Company (active) & 25,017 & 79.0\% & 52.3\%  \\
Company (balanced) & 23,812 & 80.0\% & 71.0\%  \\
\bottomrule
\end{tabular}
}
}
\end{table}

\paragraph{Finetuning}
Recent work has shown promising results using unsupervised language modeling, followed by transfer learning to natural language tasks \protect\cite{Radford2017}, \protect\cite{Radford2018}. Furthermore, these models benefit when the entire model is fine-tuned on the transfer task, as demonstrated in \protect\cite{Howard2018}. 

Specifically, these methods have beaten the state of the art on binary sentiment classification. These models have also attained the best overall score on the GLUE Benchmark\footnote{https://gluebenchmark.com/leaderboard} \protect\cite{GLUE2018}, comprised of a variety of text understanding tasks, including entailment and question answering.

\section{Methodology}
We use a larger batch size with shorter sequence length, specifically a global batch of 512 and sequence length 64 tokens (tokenized with a 32,000 BPE vocabulary, as detailed in \textbf{Characters and Subword Units}. The shorter sequence length works well because the transfer target are tweets, which are short pieces of text.  We trained our language model on the Amazon Reviews dataset \protect\cite{McAuley2015} rather than other large datasets like BooksCorpus \protect\cite{BooksCorpus2015}, because reviews are rich in emotional context. 

We also train an mLSTM network on the same dataset, based on the model from \protect\cite{Puri2018}. 

We chose to compare these particular models because they work in fundamentally different ways and because they collectively hold state of the art results on many significant academic NLP benchmarks. We wanted to test these models on difficult classification problems with real-world data.

\paragraph{Unsupervised Pretraining.} 
The language modeling objective can be summarized as a maximum likelihood estimation problem for a sequence of tokens. We treat our model as a function with two parts: an encoder $f_e$ and decoder $f_d$. The encoder forms the bulk of the model, including the token embedding dictionary as the first module. The decoder is simply a softmax linear layer that projects the encoder output into the dimension equal to the vocabulary size. The objective to maximize is as follows. 
\begin{align*}
-\log{p(x_0,\ldots, x_n)} &= -\sum_{t=1}^n \log{p(x_t | x_{t-1}, \ldots, x_0)} \\
p(x_t | x_{t-1}, \ldots, x_0) &= f_d(h^l_t)
\end{align*}
where $h^l_t$ is a hidden layer activation in the final layer of $f_e$, indexed $1 \ldots l$ for timestep $t$. 

The model is tasked with predicting the next token given all of the ones prior by outputting a probability distribution over the vocabulary of tokens. Doing this for each timestep $t$ produces each term in the sum of the log-likelihood formulation, and so maximizing the correct probabilities is a way to understand the joint probability distribution of sequences in this corpus of text.

\paragraph{Characters and Subword Units.}
While \protect\cite{Radford2017}, \protect\cite{Gray2017} and \cite{Puri2018} have shown state of the art results for language modeling and task transfer with character-level mLSTM models, we found that our Transformer model benefits from modeling language through subword units. Using a byte-pair-encoding (BPE) \protect\cite{BPE2015} of various sized %(Fig.\ref{table:BPE-sentencepiece}), 
we notice that a 32,000 word-piece vocabulary achieves a better bits per character (BPC) loss over one epoch of the Amazon Reviews dataset \protect\cite{McAuley2015} than a small vocabulary. We compute the BPC equivalent over word pieces, following \protect\cite{SubwordMikolov2012}. 
For the remainder of this work, our Transformer models use 32,000 word pieces\footnote{Library for BPE available in open source: https://github.com/google/sentencepiece}. 

Recent work \protect\cite{DeepXformer2018} has shown that it is possible to train a character level Transformer that is up to 64 layers deep and which beat state of the art BPC over large text datasets. However this requires intermediate layer losses, and other auxiliary losses for optimal convergence. By comparison, \protect\cite{Radford2018} uses a bytepair encoding vocabulary with 40,000 word pieces for their state of the art results on language transfer tasks with a Transformer model. Our work closely follows their model.

\paragraph{Supervised Finetuning.}
After the pretraining, we initialize a new decoder $f_d^{\dagger}$ to be exclusively trained on the supervised problem. Depending on the task, this decoder may be a single linear layer with activation or an MLP. We also retain the original decoder $f_d$ and continue to train it by using language modeling as an auxiliary loss when finetuning on the new corpus. Error signals from both decoders are backpropagated into the language model. The differences between the hyperparameters for finetuning and language modeling are described in Table \ref{table:hparams}.

\begin{table}[!t]
\caption{Hyperparameters for language modeling and finetuning phases.}
\label{table:hparams}
\makebox[\columnwidth][l]{
\resizebox{\columnwidth}{!}{

\begin{tabular}{p{1.5cm}|p{4cm}p{4cm}}
\toprule
 & Language Modeling & Finetuning \\
\midrule
global batch size & 512 (size 64 on 8 GPUs) & 32 \\ 
\midrule
sequence length & 64 - kept short because targeting tweet application& $\max(\text{batch})$ \\
\midrule
optimizer & \multicolumn{2}{c}{ADAM} \\
\midrule
$lr$ \newline (schedule) & $2 \times 10^{-4}$ \newline (cosine decay after linear warmup on 2000 iterations) & $1 \times 10^{-5}$ \newline (constant after 1/2 epoch linear warmup) \\
\midrule
Decoder module & $\mathbb{R}^{d_h \times 32000}$ & \textit{Binary:} MLP(1024 $\rightarrow$ $n_c$) with PReLU and 0.3 dropout \newline 
\textit{Multiclass:} MLP(4096 $\rightarrow$ 2048 $\rightarrow$ 1024 $\rightarrow$ $n_c$) with PReLU and 0.3 dropout \\
\midrule
\# Epochs & 1  & 5 \\
\midrule
Loss & $\mathcal{L}_{LM}$ = Softmax Cross Entropy & Sigmoid Binary Cross Entropy $+ 0.02 \cdot \mathcal{L}_{LM}$ \\

\bottomrule
\end{tabular}

}
}
\end{table}

\paragraph{ELMo Baseline}
We also compare our language models to ELMo \protect\cite{ELMo}, a contextualized word representation based on a deep bidirectional language model, trained on large text corpus. We use a publicly available pretrained ELMo model from the authors. During finetuning, text is embedded with ELMo before being passed into a decoder $f_d^{\dagger}$. Error signals are backpropagated into the ELMo language model. Unlike our other models, we do not use an auxiliary language modeling loss during finetuning, as the ELMo language model is bidirectional.

Finetuning the ELMo model substantially improves accuracy on our tasks, thus we include only finetuned ELMo results. 

\paragraph{Multihead vs. Single Head Finetuning Decoders}
The tweet datasets are an example of a multilabel classification problem. We can formulate the problem for the finetuning decoder, $f_d^{\dagger}$ as either a collection of single binary problems or multiple problems put together. 

The single binary problem formulation allows for a focus on one class and end-to-end optimization will only have one error signal. However, because the label classes are imbalanced in all categories, this may lead to a sparse gradient signal for the positive label, which may impact recall and precision. Increasing the size of $f_d^{\dagger}$ to more than one linear layer leads to rapid overfitting and lower validation performance.

The combined binary problems formulation (henceforth described as \textit{multihead}) allows for a richer error signal that propagates more information through the encoder $f_e$ and sentiment representation in $f^{\dagger}_d$. In this setup, constructing a Multilayer network is far more useful, and can be thought of as specifically creating sentiment features to be used at the final layer to predict the presence of the individual emotions. We find that the inclusion of easier, more balanced label categories improves performance on harder ones in Table \ref{table:transformer-vs-mlstm}. However, the easier categories have slightly lower performance because the network is not being optimized for only those categories. 

\paragraph{Thresholding Supervised Results}
For both the multihead MLP and the single linear layer instantiating of $f_d^{\dagger}$, we found that thresholding predictions produced noticeably better results than using a fixed threshold value such as $t^* = 0.5$. This makes sense since the label classes for most categories are very imbalanced. For thresholding, we take a dataset of tweets and split it into training (70\%), thresholding (10\%) and validation (20\%) sets. At each epoch of finetuning on the training set, we calculate validation accuracy and save predictions on the threshold set on the epoch for which this is maximized. 

To threshold, we search the discretized version of \text{[0, 1]}: the linear space $T = \{ \frac{i}{200}: \, \: 1 \le i \le 200 \}$ for the positive label threshold for each category. We denoted the threshold which gave the best score on the threshold set as $t^*$. 

IBM Watson and Google NLP\footnote{https://cloud.google.com/natural-language/} both offer commercial APIs for binary sentiment analysis, producing scalar values that correspond to a continuous [-1,+1] sentiment score. We applied our thresholding procedure to these scores. In the case of classification with neutrals we create two thresholds $0 < t_1^* < t_2^* < 1$ which we individually optimized jointly over $T$ as well. With the finetuning procedure, we found success with a decoder $f_d^{\dagger} = \text{MLP}(64, 2)$, whose two output units $\hat{y}_p, \hat{y}_n$ are probability estimates of the positive and negative labels $y_p, y_n$. These units both have sigmoid activations, since we denote a neutral as $y_p = y_n = 0$. To threshold these predictions, we searched the cartesian product $T \times T$ to determine $0 < t_p^*, t_n^* < 1$. 

\paragraph{Active Learning}
We hypothesized that we could achieve greater precision and recall on our datasets if our class label were more equally balanced. To this end, we employed an active learning procedure to select unlabeled tweets to be labeled. The algorithm consisted of first finetuning a language model $f = (f_e, f_d, f_d^{\dagger})$ on labeled tweets for 5 epochs. At peak validation accuracy, we obtain predictions $P \in \mathbb{R}^{8 \times n_{u}}$, for Plutchik sentiment on the unlabeled tweets. 

From the labeled dataset, we calculate the negative class percentage for each category $v \in \mathbb{R}^{8}$. Then we obtain category a weighting parameter $w = 10 \times (v - 0.5)$ so that $w_i \in [-5, 5]$ for $i \in 1\ldots8.$ Then, we get scores for each unlabeled point as weighted features: $s = e^{w^{\top}P} \in \mathbb{R}^{n_u}$. This way, positive predictions for sentiment categories are weighted by how much they would contribute towards balancing all of the class distributions. The scores $s$ are used as weights in a weighted uniform random sampler, and from this, we sampled 5,000 tweets to be labeled. 

We found that overall, the method produced tweets with more emotion. Not only was the positive class balance averaged across label categories higher (11.2\% compared to 8.2\% for random sampling),  but the percentage of tweets which had no emotion was dramatically lower: 35.6\% compared to 52.1\% for random sampling (Table \ref{table:class-imbalance}). We hence achieved better class balance than the dataset prior to the augmentation. 

% Binary Results table
\begin{table*}[!htpb]
\centering
\caption{Binary sentiment accuracy. The SST dataset includes Positive and Negative labels. Other datasets include Neutral labels. Third party results (Watson and Google) thresholded on the test set. }
\label{table:binary-results}
\makebox[\textwidth]{
\resizebox{0.67\textwidth}{!}{%
\begin{tabular}{cccc}
\toprule
& SST (acc) & Company & -/=/+ \\ % & SemEval & +/=/- \\
\midrule
Transformer (finetune) & 90.9\% & \textbf{81.2\%} & \textbf{88.2}/\textbf{73.5}/\textbf{81.9}  \\
%mLSTM (no finetune) & 91.5\% & 49.8\% & 42.8/52.1/54.7 & -  \\
mLSTM (finetune) & 90.4\% & 78.2\% & 87.0/69.3/78.3  \\
8k mLSTM\cite{Puri2018} & \textbf{93.8\%} & 77.3\% & 86.0/67.4/78.6  \\
\cite{Gray2017} & 93.1\% & - & -   \\
ELMo (finetuned) & 79.9\% & 71.4\% & 81.7/60.1/72.4 \\
ELMo+BiLSTM+Attn \cite{GLUE2018} & 91.6\% & - & - \\
Watson API & 84.4\% & 56.7\% & 42.9/54.0/73.3  \\
Google Sentiment (GCL) API & 81.3\% & 62.5\% & 69.6/54.0/63.8 \\
\midrule
Class Balance & 50.0/50.0 & - & 22.4/46.0/31.6 \\
\bottomrule
\end{tabular}%
}
}
\end{table*}

\begin{table}[!hpb]
%\begin{table*}[!htpb]
% TODO: Citation and link for SemEval competition 
\caption{Comparison on SemEval Task 1:E-c challenge. Official results on the golden test set [truth labels hidden]. \footnote{https://competitions.codalab.org/competitions/17751}}
\label{table:SemEval-results}
\makebox[\columnwidth][l]{
\resizebox{\columnwidth}{!}{
\begin{tabular}{cccc}
\toprule
& Accuracy & \multirow{2}{*}{Micro F1} & \multirow{2}{*}{Macro F1} \\
& (Jaccard) &   & \\
\midrule
Transformer (ours) & 0.577 & 0.690 & \textbf{0.561} \\
\cite{WinnerSemEval2018Task1} & \textbf{0.595} & \textbf{0.709} & 0.542 \\
\cite{ThirdPlaceSemEval2018Task1} & 0.582 & 0.694 & 0.534 \\
\bottomrule
\end{tabular}
}
}
\end{table}

\begin{table*}[!htpb]
\caption{Transformer vs. mLSTM on Plutchik Tweet Categories (F1 Score). MH: Multi Head, SH: Single Head}
\label{table:transformer-vs-mlstm}
\makebox[\textwidth]{
\resizebox{\textwidth}{!}{
\begin{tabular}{cccccccccc|c}
\toprule
& & Anger & Anticipation & Disgust & Fear & Joy & Sadness & Surprise & Trust & \textbf{Average}\\
\midrule
% Company tweets Multihead: (Nikolai)
% anger + 68.40, anticipation + 48.63, disgust + 44.12, fear + 40.00, joy + 63.39, sadness + 33.33, surprise + 26.89, trust + 30.00;
\multirow{3}{*}{Company} & Transformer (MH) & \textbf{.684} & .486 & \textbf{.441} & \textbf{.400} & .634 & \textbf{.333} & \textbf{.269} & \textbf{.300} & \textbf{.443} \\
% \multirow{3}{*}{Company} & Transformer (MH) & .607 & .457 & .476 & .180 & .692 & .314 & .275 & .233 & \textbf{.404} \\
% Company tweets single head each:
% & Transformer (SH) & .678 & .456 & .342 & .333 & .611 & .274 & .161 & .263 & .390 \\
% Company tweets, re-computed (Nikolai)
& Transformer (SH) &  .679 & \textbf{.491} & .371 & \textbf{.400} & \textbf{.675} & .286 & .210 & .279 & .424 \\
& mLSTM (SH) & .636 & .426 & .319 & .232 & .609 & .260 & .201 & .284 & .371 \\
& ELMo (MH) & .515 & .306 & .325 & .086 & .489 & .182 & .161 & .182 & .281 \\
& Watson & .358 & - & .179 & .086 & .520 & .096 & - & - & -   \\

\midrule
% Transformer text with no formatting, transliterated emoji (Nikolai)
% anger + 77.89, anticipation + 41.31, disgust + 76.92, fear + 72.30, joy + 85.02, sadness + 71.25, surprise + 36.00, trust + 23.96
\multirow{3}{*}{Semeval} & Transformer (MH) & \textbf{.779} & .413 & \textbf{.769} & .723 & \textbf{.850} & \textbf{.712} & .360 & .240 & \textbf{.606} \\
% Neel numbers:
%\multirow{3}{*}{Semeval} & Transformer (MH) & .735 & \textbf{.366} & \textbf{.731} & .661 & .782 & .629 & \textbf{.364} & .044 & .539 \\
% Question: Don't we get better Single Head results in Neel table?
% Neel numbers:
%& Transformer (SH) & .751 & .359 & .729 & .672 & .828 & .636 & .333 & .043 & .544 \\
% Transformer text with no formatting, transliterated emoji (Nikolai)
& Transformer (SH) & .774 & \textbf{.425} & .765 & \textbf{.735} & .832 & .699 & \textbf{.373} & \textbf{.247} & \textbf{.606} \\
& mLSTM (SH) & .668 & .189 & .691 & .535 & .763 & .557 & .103 & .000 & .438 \\
& ELMo (MH) & .506 & .215 & .351 & .172 & .540 & .348 & .164 & .239 & .317 \\
& Watson & .498  & - & .331 & .149 & .684 & .359 & - & - & - \\
% Raul numbers:
%& Transformer (SH) & .595 & \textbf{.381} & .419 & .374 & .694 & .511 & .241 & .331 & .443 \\
%& mLSTM (SH) & .552 & .285 & .418 & .250 & .676 & .484 & .184 & .293 & .393 \\
\bottomrule
\end{tabular}
}
}
\end{table*}

\section{Results}
\subsection{Binary Sentiment Tweets}
For binary sentiment, we compare our model on two tasks: the academic SST dataset, which consists of a balanced set of Positive and Negative labels, and the company tweets dataset, which consists of a balance between Positive, Neutral and Negative labels. See Table \ref{table:binary-results}.

While the Transformer gets close but does not exceed the state of the art on the SST dataset, it exceeds both the mLSTM and ELMo baseline as well as both Watson and Google Sentiment APIs on the company tweets. This is despite optimally calibrating the API results on the test set.

\subsection{Multi-Label Emotion Tweets}
The IBM Watson API offers multi-label emotion predictions for five categories: Anger, Disgust, Fear, Joy and Sadness. We compare our models to Watson on these categories for both the SemEval dataset and the company tweets in Table \ref{table:transformer-vs-mlstm}. We find that our models outperform Watson on every emotion category. 

\paragraph{SemEval Tweets}
We submitted our finetuned Transformer model to the SemEval Task1:E-C challenge, as seen in Table \ref{table:SemEval-results}. These results were computed by the organizers on a golden test set, for which we do not have access to the truth labels. Our model achieved the top macro-averaged F1 score among all submission, with competitive but lower scores for the micro-average F1 an the Jaccard Index accuracy \footnote{SemEval 2018 results can be seen at 
http://alt.qcri.org/semeval2018/. Our entry is $\#1$ in the post-evaluation period for Task1:E-C, as of October 2018.}. This suggests that our model out-performs the other top submission on rare and difficult categories, since macro-average weighs performance on all classes equally, and the most common categories of Joy, Anger, Disgust and Optimism get relatively higher F1 scores across all models. 

We also compare the deep learning architectures of the Transformer and mLSTM on this dataset in Table \ref{table:transformer-vs-mlstm} and find that the Transformer outperforms the mLSTM across Plutchik categories.

The winner of the Task1:E-c challenge \cite{WinnerSemEval2018Task1} trained a bidirectional LSTM with an 800,000 word embedding vocabulary derived from training word vectors \cite{Mikolov2013} on a dataset of 550 million tweets. Similarly, the second place winner of the SemEval leaderboard trained a word-level bidirectional LSTM with attention, as well as including non-deep learning features into their ensemble \cite{ThirdPlaceSemEval2018Task1}. Both submissions used training data across SemEval tasks, as well as additional training data outside of the training set.

In comparison, we demonstrate that finetuning can be as effective on this task, despite training only on 7,000 tweets. Furthermore, out language modeling took place on the Amazon Reviews dataset, which does not contain emoji, hashtags or usernames. We would expect to see improvements if our unsupervised dataset contained emoji, for example.

\paragraph{Plutchik on Company Tweets}
Our models gets lower F1 scores on  the company tweets dataset than on equivalent SemEval categories. 
As with the SemEval challenge tweets, the Transformer outperformed the mLSTM. These results are shown in Table \ref{table:transformer-vs-mlstm}. Both models performed significantly better than the Watson API on all categories for which Watson supplies predictions.

We could not conclusively determine whether the singlehead or the multihead Transformer will perform better on a given task. Thus we recommend trying both methods on a new dataset.

\section{Analysis}

\paragraph{Classification Performance by Dataset Size}
We would have liked to label more data for the company tweets dataset, and thus looked into how much extra labeling contributes to finetuned model performance accuracy. 

First, let us explain the difference between micro and macro averaging of the F1 scores. We can summarize the F1 scores of categories $c \in \mathcal{C}$ (or any other metric $M$) through macro and micro averaging to obtain $\overline{M}$. The macro method weights each class equally by averaging the metric calculated on each individual class. The micro method accounts for the class imbalances in each category by aggregating all of the true/false positives/negatives first, and then calculating an overall metric. 
\begin{align*}
\overline{M}_{\text{macro}} &= \frac{1}{|\mathcal{C}|} \sum_{c \: \in \: \mathcal{C}} M(TP_c, TN_c, FP_c, FN_c) \\
\overline{M}_{\text{micro}} &= M(\overline{TP}, \overline{TN}, \overline{FP}, \overline{FN}) \\
\overline{TP} &= \sum_{c \: \in \: \mathcal{C}} TP_c, \,  \overline{TN} = \sum_{c \: \in \: \mathcal{C}} TN_c\ldots
\end{align*}

\begin{figure*}[htpb!]
    % \makebox{\textwidth}{
    % \resizebox{\textwidth}{!}{
    \centering
    \begin{subfigure}[t]{0.263\textwidth}
        \centering
        \caption{\label{fig:macro-micro}}
        \includegraphics[width=\textwidth]{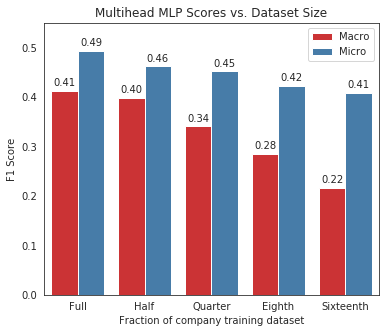}
        % \caption{Comparison of macro and micro averages of F1 scores across categories on the company tweets dataset.}
    \end{subfigure}
    ~ 
    \begin{subfigure}[t]{0.717\textwidth}
        \centering
        \caption{\label{fig:cat-sizes}}
        \includegraphics[width=\textwidth]{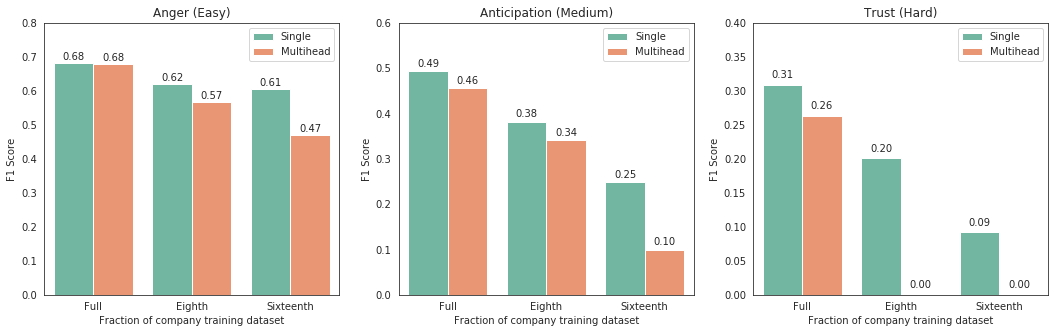}
        % \caption{F1 Scores for different categories on different dataset sizes for single head vs. multihead decoder}
    \end{subfigure}%
    \caption{\subref{fig:macro-micro}) Comparison of macro and micro averages of F1 scores across categories on the company tweets dataset. \subref{fig:cat-sizes}) F1 Scores for different categories on different dataset sizes for single head vs. multihead decoder.}
    \label{fig:data-fractions}
    % }
    % }
\end{figure*}

In one experiment, we decreased the size of the training dataset and observed the resulting macro and micro averaged F1 scores across all categories on company tweets. The results are shown in Fig. \ref{fig:macro-micro}. We observe that the macro average is more sensitive to dataset size and falls more quickly than the micro average. The interpretation of this is that categories with worse class imbalance (which consequently influence macro more than micro average) benefit more from having a larger training dataset size. This suggests that we may obtain substantially better results with more data in the harder categories.

We conducted a related experiment that focused on the difference in category performance when using a single head versus a multihead decoder $f_d^{\dagger}$. We apply the two architectures at different training dataset sizes  for three different label categories: Anger, Anticipation and Trust, which we categorize as \textit{low, medium} and \textit{high} difficulty, respectively. As seen in Fig. \ref{fig:cat-sizes} it appears that the difference between the single and multihead becomes more pronounced for more difficult categories, as well as for smaller dataset sizes. 

We do not have enough data to make a firm conclusion, but this study suggests that we could get more out of the labeled data that we have, by studying which categories benefit from single head and multihead decoders. All categories benefit from more training data, but some categories benefit from from marginal labeled data than others. This suggests further and more rigorous study of the boostrapping methods we used to select tweets for our human labeling budget, as described in the \textbf{Active Learning} section.

\paragraph{Dataset Quality and Human Rater Agreement}
The SemEval dataset \cite{SemEval2018Task1} applies a positive label for every category where 2 out of 7 vetted raters agree. The reason is for the dataset to contain difficult and subtle examples of sentiments, not just those examples where everyone agrees. The raters also have a tendency to under-label categories, especially when presented multiple options. 

Following a similar process, we required 2 out of 5 raters for a positive label, and in the case of binary sentiment labels (Positive, Neutral, Negative), we rounded toward polarized sentiment and away from Neutral labels in the case of a 2/3 split. Applying the SemEval-trained Transformer directly to our company tweets dataset gets reasonably good results (0.338 macro average), also validating that our labeling technique is similar to that of SemEval.   

Looking at rater agreement by dataset (Fig. \ref{table:rater-agreement}), we see that Plutchik category labels contain large rater disagreement, even among vetted raters who passed the golden set test. Furthermore, datasets with more emotions (the SemEval dataset and our active learning sampled company tweets) contain higher Plutchik disagreement than random company tweets. This is likely because raters tend to apply the "No Emotion" label when they are not sure about a category. As Table \ref{table:class-imbalance} shows, the SemEval and active company tweets datasets contain fewer no-emotion tweets than other datsets.

It would be interesting to analyze rater disagreement by category, how much this effects classifier convergence, whether getting 7+ ratings per tweet helps classifier convergence, and also whether this work could benefit from estimating rater quality via agreement with the crowd, as proposed in \cite{NoisyLabelHumanAWS2017}. However this analysis is not straightforward, as the truth data is itself collected through human labeling. 

Alongside classifier convergence by dataset size (Fig.\ref{fig:cat-sizes}), we think that this could be an interesting area a future research.

\paragraph{Difficult tweets and challenging contexts.}
There is not sufficient space for a thorough analysis, but we wanted to suggest why general purpose APIs may not work well on our company tweets dataset. Table \ref{table:video-game-tweets} samples the largest binary sentiment disagreements between human raters and the Watson API. For simplicity, we restrict examples to video game tweets, which comprise 19.1\% of our test set. As we can see, all of these examples appear to ascribe negative emotion to generally negative terms which, in a video game context, do not indicate negative sentiment.

Our purpose is not to castigate the Watson or the GCL APIs. Rather, we propose that it may not be possible to provide context-independent emotion classification scores that work well across text contexts. 

It may work better in practice, on some tasks, to train a large unsupervised model and to use a small amount of labeled data to finetune on the context present in the specific dataset. We would like to quantify this further in future work. 

Recent work \cite{SoftmaxBottleneck2017} shows that training an RNN with multiple softmax outputs leads to a much improved BPC on language modeling, especially for diverse datasets and models with large vocabularies. This is because the multiple softmaxes are able to capture a larger number of distinct contexts in the text than a single output. 

Perhaps our Transformer also captures the features relevant to a large number of distinct contexts, and the finetuning is able to select the most significant of these features, while ignoring those features that -- while adding value in general -- are not appropriate in a video game setting.

\section{Conclusion}
In this work we demonstrate that unsupervised pretraining and finetuning provides a flexible framework that is effective for difficult text classification tasks. 
We noticed that the finetuning was especially effective with the Transformer network, when transferring to downstream tasks with noisy labels and specialized context.

We think that this framework makes it easy to customize a text classification model on niche tasks. Unsupervised language modeling can be done on general text datasets, and requires no labels. Meanwhile downstream task transfer works well enough, even on small amounts of domain-specific labelled data, to be accessible to most academics and small organization. 

It would be great to see this approach applied to a variety of practical text classification problems, much as \cite{Radford2018} and \cite{devlin2018bert} have applied language modeling and transfer to a variety of academic text understanding problems on the GLUE Benchmark.

\bibliography{aaai}
\bibliographystyle{aaai}

\end{document}